%% file: acl_latex.tex
\pdfoutput=1

\documentclass[11pt]{article}

\usepackage[]{acl}

\usepackage{times}
\usepackage{latexsym}
\usepackage{graphicx}
\usepackage[T1]{fontenc}
\usepackage{algorithm}
\usepackage{algorithmic}
\usepackage[utf8]{inputenc}

\usepackage{microtype}

\usepackage{newfloat}
\usepackage{listings}
\usepackage{subcaption}
\floatstyle{ruled}
\newfloat{listing}{tb}{lst}{}
\floatname{listing}{Listing}
\newcommand{\TaBERT}{{\sc TaBert}}
\newcommand{\TAPAS}{{\sc TaPas}}
\newcommand{\RCI}{{\sc RCI}}

\title{\projName: End-to-End Table Question Answering via Retrieval-Augmented Generation}
\author{
{Feifei Pan$^{1}$}, {Mustafa Canim $^{2}$}, {Michael Glass $^{2}$}, {Alfio Gliozzo$^{2}$}, {James Hendler$^{1}$}\\
{\tt panf2@rpi.edu}, {\tt mustafa@us.ibm.com}, \\{\tt mrglass@us.ibm.com}, {\tt gliozzo@us.ibm.com}\\{\tt hendler@cs.rpi.edu}
\\
$^{1}$ Rensselaer Polytechnic Institute \\
$^{2}$ IBM TJ Watson Research Center {\tt } \\
 \\
}

\date{}
\input{macros}

\begin{document}

\maketitle

\input{0abstract}
\input{1intro}
\input{2related}
\input{3model}
\input{4experiments}
\input{6conclusion}

\bibliography{acl22,custom}
\bibliographystyle{acl_natbib}




\end{document}

%% file: macros.tex
\usepackage{xcolor}
\newcommand{\projName}{T-RAG}
\newcommand{\task}{Table QA}

\newcommand{\eat}[1]{}

%% file: 0abstract.tex
\begin{abstract}

Most existing end-to-end Table Question Answering (Table QA) models consist of a two-stage framework with a retriever to select relevant table candidates from a corpus and a reader to locate the correct answers from table candidates. Even though the accuracy of the reader models is significantly improved with the recent transformer-based approaches, the overall performance of such frameworks still suffers from the poor accuracy of using traditional information retrieval techniques as retrievers. 
To alleviate this problem, we introduce \projName, an end-to-end Table QA model, where a non-parametric dense vector index is fine-tuned jointly with BART, a parametric sequence-to-sequence model to generate answer tokens. Given any natural language question, \projName\ utilizes a unified pipeline to automatically search through a table corpus to directly locate the correct answer from table cell.
We apply \projName\ on recent open-domain \task\ benchmarks and demonstrate that the fine-tuned \projName\ model is able to achieve state-of-the-art performance in both the end-to-end \task\ and the table retrieval tasks.

\end{abstract}

%% file: 1intro.tex
\section{Introduction}
Tabular data is commonly seen in open-domain documents ~\cite{cafarella2009data, Zhang2018AdHT}, such as the Web and Wikipedia, as well as in domain-specific papers, journals, manuals, and reports. Answering questions over these tables requires table retrieval and understanding of the table structure and content. \task\ task is generally more challenging than executing SQL queries over relational database tables due to the lack of schema information. 
Most existing studies tackle \task\ as two separate sub-tasks: (1) Table retrieval~\cite{webtables, cafarella2009data, Zhang2018AdHT, shraga2020ad, sigir20}, and (2) QA over tables~\cite{yu2019spider, tapas,tabert,rci_inreview}. Recently, the DTR~\cite{herzig2021open} and the CLTR~\cite{pan-etal-2021-cltr} models have been proposed as end-to-end solutions for \task. Both models consist of a two-step pipeline of a retriever to generate a set of candidate tables and a reader to answer questions over these tables. The two components are trained individually, causing error propagation from retrievers to readers, i.e. with incorrect table candidates, it is impossible for the readers to locate the correct answer despite the design of the models. While dense retrieval and Retrieval Augmented Generation (RAG)~\cite{dpr, rag} have achieved great success in open-domain QA over free text, none of the studies in the literature leverage a non-parametric memory model along with a parametric memory model for the open-domain \task\ task. 

In this paper, we describe a novel end-to-end \task\ model, \projName, replacing the two-step framework with a single training process. To train \projName, we utilize Dense Passage Retrieval (DPR) ~\cite{dpr} and RAG strategies. Specifically, we jointly train a DPR component ~\cite{glass-etal-2021-robust} together with the BART-based~\cite{bart} sequence-to-sequence (Seq2Seq) model. To the best of our knowledge, \projName\ is the first \task\ model where the query encoder for a non-parametric dense vector index is fine-tuned along with a parametric generation model. We evaluate the performance of \projName\ on NQ-TABLES~\cite{herzig2021open} and E2E\_WTQ~\cite{pan-etal-2021-cltr}, two recent end-to-end \task\ benchmarks. The experimental results indicate that \projName\ outperforms the state-of-the-art models on the end-to-end \task\ task.

The major contribution of this work is that, we propose the first end-to-end \task\ pipeline, leveraging DPR along with the Seq2Seq component of RAG. \projName\ employs a simple but effective one-step training that reduces error accumulations and simplifies model fine-tuning. In the experiments, \projName\ achieves state-of-the-art performance on two tasks. We find \projName\ improves the results for end-to-end \task\ on two recent benchmarks. The RAG component of the end-to-end model fine-tuned over \task\ benchmarks also yields state-of-the-art results on the table retrieval task. 
    

%% file: 2related.tex
\section{Related Work}
\paragraph{Table Retrieval} Traditional table retrieval models usually concatenate tables into documents while disregarding the underlying tabular structure ~\cite{pyreddy1997tintin, wangandhu, liu2007tableseer, webtables, cafarella2009data}. New approaches are proposed to retrieve tables with a set of features of the table, query and table-query pair~\cite{zhang2018ad, yan2017contentbased, Bhagavatula2013MethodsFE, shraga2020ad}. \citet{zhang2018ad} uses semantic similarities to build an ad-hoc table retrieval model with various features. A neural ranking model is introduced in \citet{sigir20}, where tables are defined as multi-modal objects and the Gated Multi-modal Units are used to learn the representation of query-table pairs. \citet{pan-etal-2021-cltr} later follows this work and improves the table retrieval with a 2-step retriever. \citet{kostic2021multimodal} discusses the use of dense vector embeddings to enhance the performance of bi- and tri-encoder in retrieving both table and text. 

\paragraph{Table QA}
Most early \task\ solutions are fully supervised models, focusing on converting natural language questions into SQL format and using the SQL-format questions to query the given tables, as seen in \citet{yu2019spider, lin2019grammarbased, xu2018sqlnet}. 
Open-domain QA over text~\cite{yu2020survey} usually utilizes multiple knowledge sources. For instance, ~\citet{oguz2021unikqa} proposes a model can convert structured, unstructured and semi-structured knowledge into text for open-domain QA. Therefore, more recent efforts have been put into investigating the use of external knowledge in enhancing the performance of Table QA. \citet{JimenezRuizHEC20} first proposes the Semantic Web Challenge on Tabular Data to Knowledge Graph Matching (SemTab) to encourage such solutions for both table understanding and \task. 
Recently, the transformer-based, weakly supervised solutions have been proposed for \task. These solutions fall into two categories: (1) Logic form-based solution, such as the \TaBERT~\cite{tabert} model, which is trained to capture the representation of natural language sentences as well as tabular data; (2) Answer cell prediction solutions, such as \TAPAS~\cite{tapas} and the \RCI~\cite{rci_inreview} model. The current state-of-the-art \RCI\ model exploits a transformer-based framework. Instead of retrieving the table cells directly for any given question-table pairs, the \RCI\ model identifies the most relevant columns and rows independently and locates the intersection table cells as the final answers. 

\paragraph{End-to-End Table QA} 
\citet{sun2016table} publishes the first end-to-end table cell search framework. This work leverages the semantic relations between cells and maps queries to table cells with relational chains. The DTR model~\cite{herzig2021open} addresses the end-to-end \task\ problem with a table retriever and a \TAPAS-based reader model. Later, the CLTR model~\cite{pan-etal-2021-cltr} introduces a similar two-step solution, using BM25 as the retriever. The model re-ranks the BM25 results and locates the table cells using the RCI scores. Recently, \citet{chen2021open} proposes a new task for QA over both free text and tables and provides a solution including a retriever with early fusion techniques and a cross-block reader. In addition, the open-domain OTT-QA benchmark is released to evaluate models for end-to-end QA over text and table.


%% file: 3model.tex
\section{The End-to-End Table QA with \projName}
\begin{figure*}
\centering 
\includegraphics[width=\linewidth]{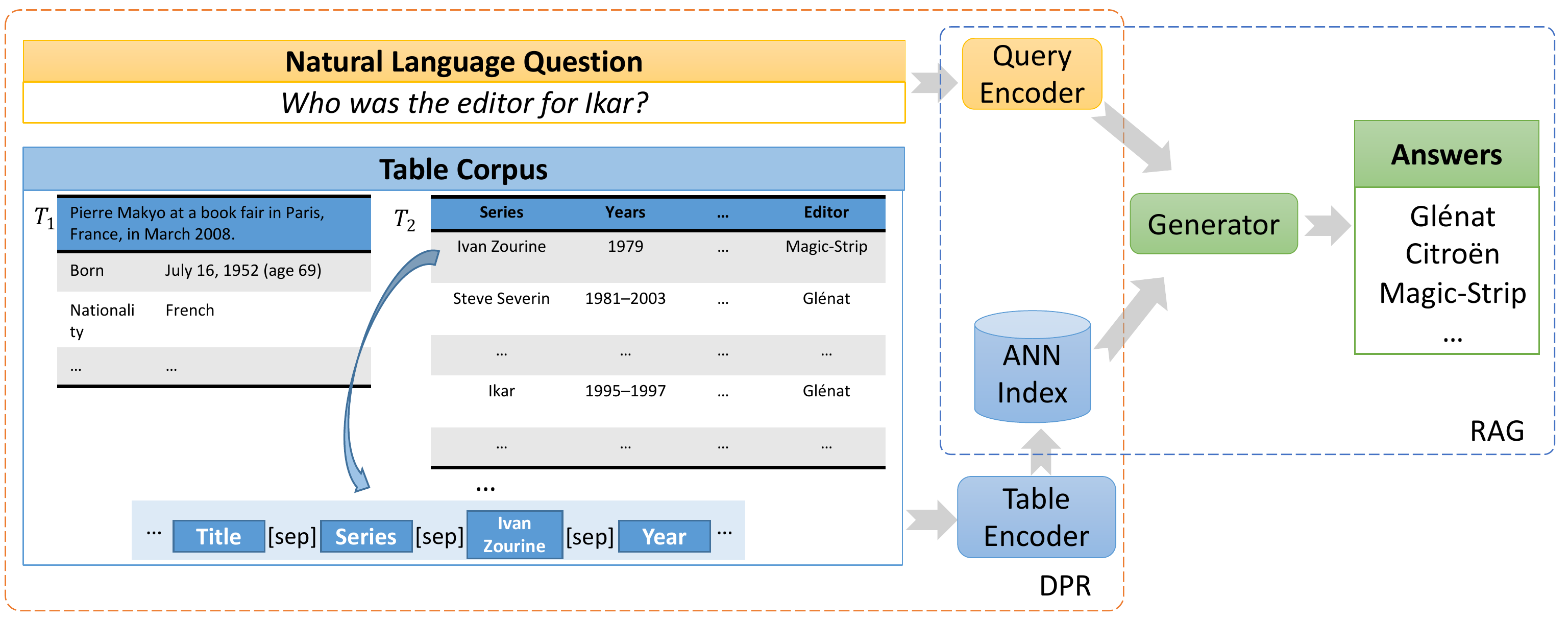}
\caption{An overview of \projName, a model trained end-to-end to directly locate answers from table corpus.} 
\label{fig.overall}
\vspace{-2 mm}
\end{figure*}

The overall architecture of \projName\ is illustrated in Figure~\ref{fig.overall}. In this example, we encode the questions ``who was the editor for Ikar?'' using the query encoder and pre-process the tables, e.g., T$_{1}$ and T$_{2}$, from the table corpus for encoding. The encoded tables are later indexed into the Approximate Nearest Neighbors (ANN) data structure for querying. The encoded question is appended to each table before inputting it to the BART-based generator for answer prediction. The DPR and the RAG components are trained jointly without explicitly considering the table-level ground truth. 
\paragraph{Setup}
We define the one-step, end-to-end \task\ task as follows. Given a massive corpus $C$ of tables $t_{i}$ and any natural language question $q_{i}$, we train a model to directly generate answer to $q_{i}$ from the table cell without any intermediate steps. Labeled datasets are available to us with ground truth samples in the format of $\{q_{i}, t_{i}, a_{i}\}$ where $a_{i}$ stands for the answers.

Table pre-processing is implemented before the training. We process the tables $t_{i}$ into a structure-preserving format, where: (1) column headers are appended before cell values, separated by a special symbol ``$\mid$''; and (2) the separator ``*'' is appended to the end of each row; (3) for the tables with additional information such as titles, we append them in front of the tables. The tables are segmented into the length of 512 tokens for training. For each question, we retrieve hard negatives from the corpus $C$ and use them as additional negative samples to enhance the \projName\ training. 
\paragraph{Soft Hard Negatives}
We implement a BM25-based hard negative mining for \projName. For each question, we first retrieve a pool of the most relevant tables from the corpus using BM25. From the table pool, we discard the ground truth table. The top-ranked, non-positive tables are used as the hard negative candidates. In the training process, instead of using the top 1 negative table, we exploit a soft hard negative mechanism, where we select the hard negative at random from the top $k$ negative tables. 

\paragraph{RAG} 
For the implementation of RAG, we jointly train a DPR-based retriever and a BART-based generator. We index the tables in $C$ using a keyword-based search engine, Anserini\footnote{https://github.com/castorini/anserini}, to harvest the hard negative training samples using BM25. Later, \projName\ exploits BERT$_{BASE}$ to encode questions along with the ground truth table and the hard negative tables. 
To train RAG, \projName\ employs the answer-level ground truth and use a Seq2Seq generator, the BART$_{LARGE}$ model, for answer predictions. The previously encoded tables are indexed with the open-source FAISS~\cite{JDH17} library into the ANN data structure for querying. The encoded questions are concatenated to each of the top retrieved tables and used as a prompt to generate the answer. More concretely, the generator predicts probability distributions for possible answer candidates as the next token. The probability distributions are later marginalized to produce a single weighted sequence probability for each answer candidate. Finally, a standard beam search decoder~\cite{sutskever2014sequence} is used to identify the most relevant candidates as the final answers to the questions at test time. Along with the answers, our model can also return the relevant table $t_{i}$ containing the correct answers from $C$ for evaluation and annotation purposes. 







%% file: 4experiments.tex
\section{Experiments}\label{sec:eval}
\paragraph{Data} We validate \projName\ on two open-domain benchmarks, \textit{NQ-TABLES} and \textit{E2E\_WTQ}. NQ-TABLES is the table subset of the Natural Questions dataset~\cite{nqdatset}, with a table corpus extracted from the English Wikipedia articles and samples in the $\{q, T, a\}$ format, where $q$, $T$, and $a$ denote question, ground truth table, and answer, respectively. E2E\_WTQ contains the look-up subset of WikiTableQuestions~\cite{pasupat2015compositional}. While a substantial amount of tables in NQ-TABLES are transposed infobox tables, the E2E\_WTQ only contains well-formatted but more complex tables. The data statistics are shown in Table~\ref{tab:data}.
\begin{table}[h]
\centering
\small
    \begin{tabular}{lllll}
        Data & Train & Dev & Test & Corpus\\
        \hline
        \hline
        NQ-TABLES & 9,594 &  1,068 & 966 & 169,898\\
        E2E\_WTQ & 851 &  124 & 241 & 2,108\\
        \end{tabular}
        \vspace{-1 mm}
    \caption{Data Statistics}
    \label{tab:data}
\end{table}

\begin{table}[h]
\centering
    \begin{subtable}[h]{.5\textwidth}
    \centering
    \small
        \begin{tabular}{lllll}
        Model & EM & F1 & Oracle EM & Oracle F1\\
        \hline\hline
        DTR+hn & 37.69 & 47.70 & 48.20 & 61.50\\
        T-RAG & \textbf{43.06} & \textbf{50.92} & \textbf{50.62} & \textbf{63.18}\\
        \end{tabular}
        \vspace{-1 mm}
    \caption{End-to-end \task\ results on the test set of NQ-TABLES.}
    \label{tab:nq_end2end}
    \end{subtable}
    \hfill
    \vspace{2 mm}
    \begin{subtable}[h]{.5\textwidth}
    \centering
    \small
        \begin{tabular}{lllll}
        Model & MRR & Hit@1 \\
        \hline\hline
        CLTR & 0.5503 & 0.4675\\
        \projName & \textbf{0.5923} & \textbf{0.5065} \\
        \end{tabular}
        \vspace{-1 mm}
    \caption{End-to-end \task\ results on the test set of E2E\_WTQ.}
    \label{tab:wtq_end2end}
    \end{subtable}
    \vspace{-3 mm}
    \caption{Experimental results on End-to-end \task.}
    \vspace{-2 mm}
    \label{tab:e2e}
\end{table}

\begin{table}[h]
\centering
    \begin{subtable}[h]{.5\textwidth}
    \centering
        \small
        \begin{tabular}{llll}
        Model & R@1 & R@10 & R@50\\
        \hline\hline
        BM25 & 16.77 & 40.06 & 58.39\\
        DTR+hn & 42.42 & 81.13 & 92.56\\
        T-RAG & \textbf{46.07} & \textbf{85.40} & \textbf{95.03}\\
        \end{tabular}
        \vspace{-1 mm}
    \caption{Table retrieval results on the test set of NQ-TABLES.}
    \label{tab:nq_tableRetrieval}
    \end{subtable}
    \hfill
    \vspace{2 mm}
    \begin{subtable}[h]{.5\textwidth}
    \centering
        \small
        \begin{tabular}{llllll}
        Model & P@5 & P@10 & N@5 & N@10 & MAP\\
        \hline\hline
        BM25 & 0.5938 &0.6587 &0.5228 &0.5356 &0.4704\\
        CLTR &  0.7437 & 0.8735 & 0.6915 & 0.7119 & 0.5971 \\
        T-RAG & \textbf{0.7806} & \textbf{0.8943}  & \textbf{0.7250} & \textbf{0.7467} & \textbf{0.6404}\\
        \end{tabular}
        \vspace{-1 mm}
    \caption{Table retrieval results on the test set of E2E\_WTQ.}
    \label{tab:wtq_tableRetrieval}
    \end{subtable}
    \vspace{-3 mm}
    \caption{Experimental results on Table Retrieval.}
    \label{tab:retrieval}
    \vspace{-3 mm}
\end{table}
\paragraph{Experimental Settings} In the experiments, we first encode the questions and tables using BERT$_{BASE}$, and later jointly train the DPR-based retriever and the Seq2Seq generator of RAG. For the experiments, we set: (1) training batch size = 128; (2) number of epochs = 2; (3) learning rate = 3e-5; and (4) gradient accumulation steps = 64. 

\paragraph{Evaluation metrics:} Following the evaluation script in SQUAD~\cite{rajpurkar-etal-2016-squad}, we evaluate end-to-end \task\ using exact match (EM) and token F1 metrics for NQ-TABLES. The accuracy for the top 1 returned answer and mean reciprocal rank (MRR) are used to measure the performance on E2E\_WTQ. We also evaluate \projName\ on the table retrieval task for a fair comparison with existing work. We utilize the original metrics in \citet{herzig2021open} and \citet{pan-etal-2021-cltr}, with recall (R) for NQ-TABLES, and precision (P), normalized discounted gain (N), and mean average precision (MAP) for E2E\_WTQ.  

\paragraph{Experimental Results}
We compare the end-to-end \task\ performance of \projName\ against the state-of-the-art DTR and CLTR models in Table~\ref{tab:e2e}. We find \projName\ yields better results than the previous best models for both datasets with all evaluation metrics. 

To further validate \projName\ against the existing models, we also evaluate the model performance on table retrieval. The table retrieval results for NQ-TABLES and E2E\_WTQ are shown in Table~\ref{tab:nq_tableRetrieval} and~\ref{tab:wtq_tableRetrieval}, respectively. The results indicate that \projName\ outperforms the simple baselines models such as BM25, as well as the strong state-of-the-art models in the experiments. 

\paragraph{Qualitative Analysis} 
We further evaluate the table retrieval results on NQ-TABLES. We notice that the DPR-based baseline of our approach achieves 43.89 for R@1 and 81.57 for R@10; both outperform the state-of-the-art DTR results. In addition, the retrieval performance is further improved with the more effective end-to-end RAG training. To validate the effectiveness of our soft hard negative technique, we test the method on the E2E\_WTQ dataset. Instead of using the top 1 negative table from the BM25 results, we set $k = 3$ and achieve a 27.17\% absolute gain for Hit@1 accuracy in the end-to-end \task\ task. 

Besides, we perform thorough error analysis on on E2E\_WTQ and find that over 21\% of the errors come from questions that involve numerical values. The finding indicates that understanding different types of numbers remains a challenge in \task.

%% file: 6conclusion.tex
\section{Conclusion and Future Work}
In this paper, we present a novel \task\ model that achieves state-of-the-art performance on recent benchmarks. Instead of training a retriever and a reader model independently, \projName\ unifies the procedure into a single pipeline of only one training step, which reduces the error accumulations from two separate models. In the experiments, \projName\ outperforms the current best models for end-to-end \task. We additionally demonstrate the advantages of \projName\ with the table retrieval task, and \projName\ beats the existing numbers on both benchmarks.

In the future, we plan to validate \projName\ on domain-specific datasets, such as AIT-QA and TAT-QA ~\cite{aitqa,tatqa} and extend the model to solve multi-modal QA problems, with the corpus containing both tables and passages, as presented in the OTT-QA and Hybrid-QA benchmarks ~\cite{ottqa, hybridqa}. To further improve the model performance, we also plan to investigate algorithms to better understand numeric values. 
